# Algorithm Research of ELMo Word Embedding and Deep Learning Multimodal Transformer in Image Description

Xiaohan Cheng[1], Taiyuan Mei[2], Yun Zi[3], Qi Wang[4], Zijun Gao[5], Haowei Yang[6]

[1]Northeastern University, Boston,USA,Cheng.xiaoh@northeastern.edu

[2]Northeastern University,Boston,USA,taiyuanmei0824@gmail.com

[3]Georgia Institute of Technology,Atlanta,USA,yzi9@gatech.edu

[4]Northeastern University,Boston,USA,bjwq2019@gmail.com

[5]Northeastern University,Boston,USA,zjg.elaine@gmail.com

[6]University of Houston ,Houston,USA,hyang38@cougarnet.uh.edu

*Abstract*-Zero sample learning is an effective method for data deficiency. The existing embedded zero sample learning methods only use the known classes to construct the embedded space, so there is an overfitting of the known classes in the testing process. This project uses category semantic similarity measures to classify multiple tags. This enables it to incorporate unknown classes that have the same meaning as currently known classes into the vector space when it is built. At the same time, most of the existing zero sample learning algorithms directly use the depth features of medical images as input, and the feature extraction process does not consider semantic information. This project intends to take ELMo-MCT as the main task and obtain multiple visual features related to the original image through self-attention mechanism. In this paper, a large number of experiments are carried out on three zero-shot learning reference datasets, and the best harmonic average accuracy is obtained compared with the most advanced algorithms.

## I. Introduction

Medical image recognition technology, based on deep learning, has achieved significant breakthroughs[1-3]. However, traditional medical image classification algorithms require a large number of labeled samples and struggle to identify novel categories[4]. As new categories emerge, distinguishing them necessitates collecting numerous labeled samples and developing a new model, which is time-consuming. Zero-shot learning, which can be categorized into conventional and generalized types, offers a solution [5]. In conventional zero-shot learning, only unseen categories are present during testing, while generalized zero-shot learning, a more practical and challenging approach, involves both known and unseen categories. Current research primarily explores prototype-based and embedding-based zero-shot learning methods.

The method maps the features and quasi-semantic information of medical images to a certain vector space to achieve alignment between the two modes [6]. During the experiment, the type of sample to be detected is determined by the nearest neighbor lookup between the data to be detected and the aligned space. However, under the generalized zero-shot condition, the mapping relationship is only established for the known categories, which is likely to lead to the overfitting of the unknown categories and the obvious deviation in the prediction of the unknown categories [7]. Therefore, the core problem of the embedded method is how to make the model can extract the most typical embedded expression from the sample, and have sufficient differentiation to distinguish other types of features.

In the past, the learning method based on convolutional neural network has occupied absolute advantages in many research directions. However, with its excellent performance in medical image recognition, the current mainstream research framework has shifted from the traditional framework based on convolutional neural network to the framework based on automatic attention network based on Transformer [8-10]. This paper first studies the semantic similarity-based multi-label category loss method, fully excavates the internal correlation between known and unknown categories and solves the problem of overfitting in the general zero-shot condition of the embedded zero-shot learning method. Secondly, with ELMo-MCT as the new backbone network and the original image as the input, the features of medical images are obtained by self-attention mechanism to overcome the problem of data bias.

## II. MULTI-MODE IMAGE DESCRIPTION CONVERTER

The feature quantity is input to the encoder for self-attention learning, and the visual attention representation of the image is obtained. The next word is then predicted based on the previous work and the encoder's visual attention information [11]. The network architecture of the multimodal image description converter is shown in Figure 1

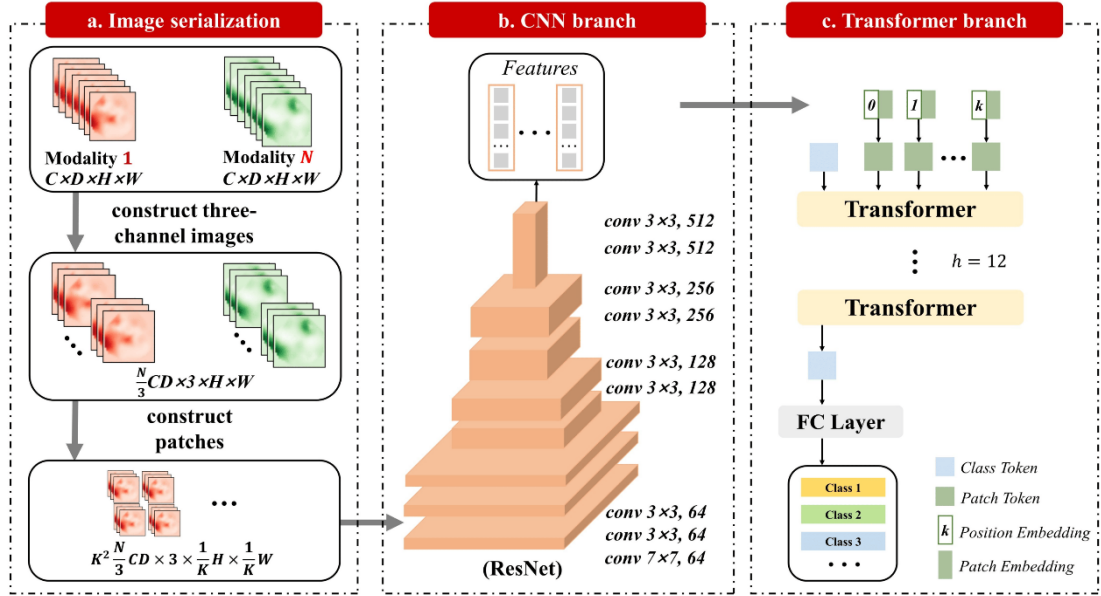

Fig. 1. Network architecture of the multimodal image description converter

### A. Image feature encoder

Its core task is to extract image characteristics and use attention mechanism to establish attention matrix between image modes, so as to obtain the correlation between images [12-14]. The algorithm mainly includes two aspects: one is the image feature extraction algorithm, the other is the multi-channel transform encoder.

#### 1) Image feature extractor

In the part of image feature extraction, the characteristics of an image are expressed [15]. This project intends to adopt the image feature extraction method based on the bottom-up attention mechanism[16], use the existing visual gene sequencing Faster-RCNN to identify the object association region[17-18], and adopt the mean pooling method to obtain the features of each object, express the characteristics of each object as $x_i$, and express the overall image as U[19]. Finally, the obtained image feature U is fed into the neural network, and the image feature matrix $U^0$ is formed by adjusting the image feature dimension to make it consistent with the spatial scale of the encoder.

#### 2) Multi-Channel Converter

The converted image feature $U^0$ is then input to a multimode converter, which contains M attention modules $\{B_e^1, B_e^2, \cdots, B_e^M\}$. The m attention module $B_e^M$ receives the output $U^{m-1}$ of the $m-1$ attention module and computes it to get the attention characteristic $U^m$ in the image mode after the attention weight. Here's the formula:

$$U^m = B_e^m(U^{m-1}) \qquad (1)$$

Each $B_e^m(U^{m-1})$ is divided into multiple concerns (MHA) and feedback forward (FFN). The multi-focus attention module consists of a single-focus attention module (Figure 2).

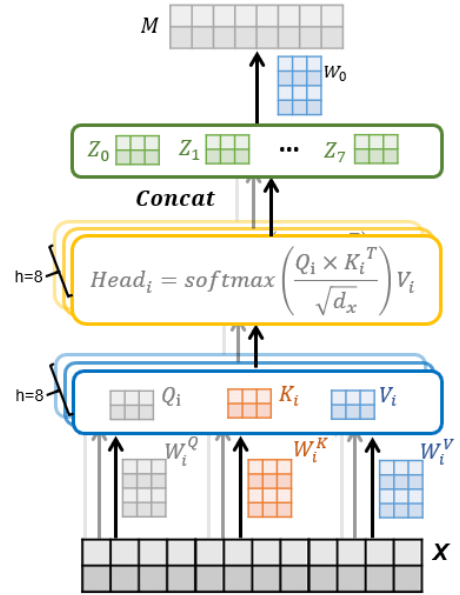

Fig. 2. Multi-head attention of the attention mechanism

Input image characteristic information $U^{m-1}$ to a single head attention module with a different parameter as shown in FIG. 2, and obtain a weight matrix $C_i^{m-1}$ representing the attention matrix between each head's image modes at layer m-1.

$$C_i^{m-1} = Attend(W,T,U) = \text{soft}\max(\frac{WT^T}{\sqrt{s}})U \quad (2)$$

$C_i^{m-1}$ is the property matrix derived from the $i$ height in level $m-1$, and $s$ is determined by the number of columns in level $W,T$. Finally, the $C_i^{m-1}$ matrix generated by the concerns is concatenated in a column.

$$\hat{C}^{m-1} = Concat(C_1^{m-1}, C_2^{m-1}, \cdots, C_l^{m-1}) \quad (3)$$

The resulting image is then linearly transformed to obtain image $C^{m-1}$.

$$C^{m-1} = Norm(U^m + G^m \hat{C}^{m-1}) \quad (4)$$

The network model is a double-layer completely connected structure with Relu as the activation function. Here's the formula:

$$FFN(C^m) = \max(0, G_1 C^{m-1} + \sigma_1)G_2 + \sigma_2 + \sigma \quad (5)$$

$$U^m = Norm(C^{m-1} + FFN(C^{m-1})) \quad (6)$$

$G, \sigma$ is the parameter of the fully connected network to be trained. By calculating the attention weight of multiple modes, the attention weight matrix $U^M$ in the multi-mode is obtained.

B. *Text decoder*

The text is decoded according to the image characteristic matrix after encoder. The specific research contents include: (1) This paper extracts the attention matrix of the correlation between words from the word pattern based on the internal attention theory of pattern; (2) Establish the association between images and text based on inter-pattern attention, obtain the image-oriented attention weight matrix, and finally generate the corresponding feature description.

*1) Multi-mode lexical information hidden encoder*

The primary task of a multimode word steganography encoder is to encode the input word and then construct the corresponding character [20]. The algorithm consists of two aspects (Figure 3 cited in Intelligent Systems with Applications, Volume 18, May 2023, 200221), namely, the standard word embedded encoder and the ELMo word embedded encoder.

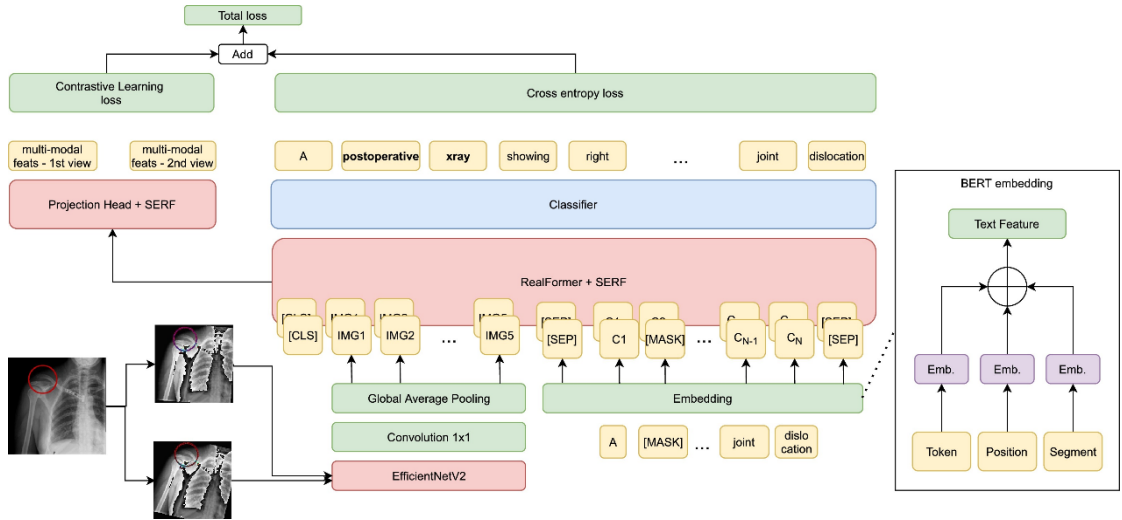

Fig. 3. Multimodal word embedding encoder

The multimodal word embedding encoder first uses the standard word embedding encoder to generate the standard word embedding[21]. The word segmentation operation is carried out, the sentence is divided into a single word, and the unique token corresponding to each word is formed, and then the input sentence is converted into the form of a token, and the completion operation is carried out [22]. The word vector encoding algorithm is then used to encode each word A, and finally B is used to represent a standard feature matrix of a description statement, where C is the dimension of the word embedding operation.

The multimodal vector encoder generates a standard vector through the standard vector encoder, that is, through segmentation, the sentence is broken down into a single word, and the corresponding symbol is established for each word, and then the sentence is converted into a symbol to complete the filling operation [23]. Then $u_i^m \in D^{emb}$ is encoded for each word-by-word vector coding, and finally $U^M$ is used to express the canonical feature matrix, where $emb$ is the dimension of lexical embedding operation.

The second step is the multimodal word embedding encoder to generate a word embedding feature matrix based on context information according to the ELMo word embedding encoder [24]. In contrast to the standard vocabulary embedding method, the ELMo vocabulary embedding encoder encodes each word in a sentence by associating it with context. The text code $U^{char}$ of each description statement is obtained through a text coding layer to avoid the occurrence of out-of-set V situation, and then the obtained text code is imported into the bidirectional LSTM to obtain the feature matrix $u^{\sigma i} \in D^{(a+1) \times G \times emb}$ with text background, where $a$ is the number of layers of LSTM and $G$ is the number of words. In this way, the feature matrix containing context is obtained. Finally, the context-containing statements are input into the mixed layer $a+1$, and the eigenmatrix is formed by weighting each vector to obtain the ELMo word vector $U^E$.

*2) Multimodal Transformer decoder*

The masked multi-head attention module models the input word embedding matrix U. Generate the top triangle first, and then do the attention model. The remaining matrix of zeros has a mask. Prevents this pattern from using words that follow the current time series. Then, the attention characteristic matrix $D_m$ in the word pattern is obtained by the attention weight of the word vector matrix.

$$D_m = Norm(U + Mask\_MHA(U,U,U)) \quad (7)$$

The image-oriented visual attention weight matrix $D_f$ is obtained by calculating the attention weight of multimodal data set $U^M$ and text data respectively. The detailed calculation method is as follows:

$$D_f = Norm(D_m + MHA(D_m, U^M, U^M)) \quad (8)$$

III. EXPERIMENTAL EVALUATION

A. *Data Set*

This project is based on Microsoft COCO2014 open-source database, and the proposed algorithm is verified by simulation [25]. The method divides the samples into 5000 image validation samples, 5000 image samples and 113287 image samples to train and test the model respectively. BLEU (A1, A2, A3, A4), ROUGE-L (R), and CIDErD (C) are used as the evaluation index and the percentage is obtained. The dataset is administered utilizing the Linked Data methodology [26], which facilitates the integration of diverse data types—a critical aspect in scholarly research. This structured approach simplifies the referencing process, thereby enhancing dataset interoperability. This characteristic proves particularly beneficial in domains such as machine learning and artificial intelligence, where the caliber of data exerts a direct influence on model training and consequently on the precision of the outcomes.

B. *Experimental Details*

Each image contains five captions in English, and the text in each caption is converted to lower case and then broken down to form a standard vocabulary about 9, 957 long. For the super parameters in this model, the feature dimension of Bottom-Up is 2048, the feature dimension of the head is 1024, the quantity h is 8, the feature dimension of the head is 128, and the term vector is 1024. Thirty cycles were trained using the cross-loss method, 10 cycles were set here, and the Adam optimizer was used for them with a learning rate of 0.0005 and a momentum of 0.9.

C. *Experimental results and analysis*

First, this project intends to take image characteristics and short-term memory (LSTM) based on Bottom-UP attention mechanism as the baseline and compare them with MCT (MCT)[27]. Secondly, this paper will make use of ELMo lexically-embedded ELMO-MCT and compare it with the above two models. The test results are shown in Table 1. Compared with the CIDEr scores of MCT and ELMo-MCT modes, the CIDEr scores of MCT and ELMO-MCT modes increased by 2.8 and 4.9 percentage points respectively.

TABLE I. ABLATION EXPERIMENT

| Model | A1 | A2 | A3 | A4 | R | C |
|---|---|---|---|---|---|---|
| BottomUP-LSTM | 78.65 | 59.06 | 46.35 | 34.27 | 57.71 | 111.04 |
| MCT | 78.75 | 61.88 | 46.77 | 35.10 | 57.92 | 113.96 |
| ELMo-MCT | 79.38 | 62.60 | 47.50 | 35.63 | 58.33 | 116.15 |

However, among the measured items, CIDEr has the largest improvement, while BLUE and ROUGE-L have a more balanced improvement. For the meaning of exponent, BLEU and ROUGE-L both judge the result by the frequency of using N lattice or longest common subsequence in candidate statements, and there is no big difference between them. CIDEr gets the final score by measuring the similarity of words. Because ELMo is used in word vector coding, the semantic meaning of the words generated by ELMO is richer, and the CIDEr score is higher than the BLEU score. Secondly, the effectiveness of the neural network under different conditions is analyzed. Tables 2 and 3 show this. The results show that the effectiveness of the method decreases with the increase of the depth. This suggests that over time, certain information may be missing, reducing the validity of the predictions.

TABLE II. COMPARISON OF MCT DEPTH TESTS FOR EACH LAYER

| Profundity | A1 | A2 | A3 | A4 | R | C |
|---|---|---|---|---|---|---|
| 2 | 78.75 | 61.88 | 46.77 | 35.10 | 57.92 | 113.96 |
| 4 | 78.65 | 61.67 | 46.67 | 35.00 | 58.13 | 114.17 |
| 6 | 77.81 | 60.73 | 45.73 | 34.27 | 57.50 | 112.08 |

TABLE III. COMPARISON OF EXPERIMENTAL RESULTS OF ELMO-MCT AT DIFFERENT DEPTHS

| Profundity | A1 | A2 | A3 | A4 | R | C |
|---|---|---|---|---|---|---|
| 2 | 79.38 | 62.60 | 47.50 | 35.63 | 58.33 | 116.15 |
| 4 | 78.96 | 61.98 | 46.77 | 35.00 | 57.92 | 114.58 |
| 6 | 78.23 | 60.94 | 45.73 | 34.06 | 57.50 | 112.50 |

Finally, the overall performance of this model can be obtained from Table 4. Through comparative analysis, it can be seen that this method is superior to most traditional models in various aspects of performance, and its CIDEr score is 2.8 and 4.9 percentage points higher than that of baseline models such as Bottom-LSTM.

TABLE IV. OVERALL PERFORMANCE OF ELMO-MCT MODEL

| Model | A1 | A2 | A3 | A4 | R | C |
|---|---|---|---|---|---|---|
| NIC | 69.38 | 48.13 | 35.83 | 25.31 | 55.21 | 89.06 |
| Soft-Attention | 73.65 | 51.25 | 35.83 | 25.31 | - | - |
| Hard-Attention | 74.79 | 52.50 | 37.19 | 26.04 | - | - |
| ATT | 76.15 | 58.85 | 44.17 | 32.92 | 55.73 | 98.23 |
| SCA-CNN | 74.17 | 56.46 | 41.77 | 31.46 | 56.46 | 95.00 |
| SCST: Att2all | - | - | - | 31.25 | 55.63 | 103.54 |
| SCST: Att2in | - | - | - | 32.60 | 56.56 | 105.52 |
| MSM | 76.98 | 59.90 | 45.42 | 34.38 | 56.46 | 104.48 |
| Bottom-UP-ResNet | 77.60 | - | - | 34.79 | 56.67 | 109.79 |
| BottomUP-LSTM (baseline) | 78.65 | 59.06 | 46.35 | 34.27 | 57.71 | 111.04 |
| MCT (ours) | 78.75 | 61.88 | 46.77 | 35.10 | 57.92 | 113.96 |
| ELMo-MCT (ours) | 79.38 | 62.60 | 47.50 | 35.63 | 58.33 | 116.15 |

## IV. CONCLUSION

Encoding it with languages such as Word2Vec and Glove resulted in the loss of a large amount of useful data. For this purpose, the image representation method of multi-mode converter is established, and the accuracy of the model is improved by joint modeling of the attention within and between modes. By introducing the method of combining ELMo and vector, the semantic expression level of the model is further improved. Overall, our findings demonstrate the potential of integrating semantic similarity measures with advanced feature extraction technologies to enhance the capabilities of zero-shot learning in medical image recognition. This strategy not only improves model performance but also reduces the dependency on extensive labeled datasets, offering a viable path forward in the development of more adaptable and efficient medical imaging technologies.